\documentclass[runningheads]{llncs}
\usepackage{graphicx}
\usepackage{comment}
\usepackage{amsmath,amssymb} 
\usepackage{color}

\usepackage{xspace}
\usepackage{color}
\usepackage{booktabs}
\usepackage{subcaption}
\usepackage{bbm}
\usepackage{microtype}
\usepackage{tabularx}
\usepackage{wrapfig}

\usepackage[T1]{fontenc}
\usepackage[utf8]{inputenc}

\newcommand{\todo}[1]{{\textcolor{red}{TODO: #1}}}

\newcommand{\eg}{e.g.\xspace}
\newcommand{\etal}{et al.\xspace}
\newcommand{\ie}{i.e.\xspace}
\newcommand{\etc}{etc.\xspace}

\newcommand{\wrt}{wrt.\xspace}
\newcommand{\vs}{vs.\xspace}

\newlength\tikzfigwidth
\newlength\tikzfigheight

\makeatletter
\renewcommand{\paragraph}{%
  \@startsection{paragraph}{4}%
  {\z@}{0.5ex \@plus 1ex \@minus .2ex}{-1em}%
  {\normalfont\normalsize\bfseries}%
}
\makeatother

\usepackage{enumitem}
\setitemize[0]{leftmargin=10pt,itemsep=0pt,topsep=2pt,parsep=0pt}
\setenumerate[0]{leftmargin=16pt,itemsep=0pt,topsep=2pt,parsep=0pt}

\def\realspace{\mathbb{R}} 
\newcommand{\cardinality}[1]{\ensuremath{\left|#1\right|}} 
\newcommand{\onehot}[1]{\ensuremath{\mathbf{1}^{#1}}} 
\newcommand{\onehote}[1]{\ensuremath{1}^{#1}} 

\def\numds{N}
\newcommand{\ds}[1]{\ensuremath{\mathtt{D}_{#1}}}
\newcommand{\dssize}[1]{\ensuremath{M_{#1}}}
\newcommand{\lblspc}[1]{\ensuremath{\mathtt{L}_{#1}}}

\def\img{I}
\newcommand{\imgi}[2]{\ensuremath{\img_{#1,#2}}}
\newcommand{\setbbgt}[2]{\ensuremath{\mathtt{G}_{#1,#2}}}
\newcommand{\bbgt}[3]{\ensuremath{g_{#1,#2}^{#3}}}

\newcommand{\setbbdt}[1]{\ensuremath{\mathtt{D}_{#1}}}
\newcommand{\bbdt}[2]{\ensuremath{d_{#1}^{#2}}}

\newcommand{\setbbpgt}[2]{\ensuremath{\hat{\mathtt{G}}_{#1,#2}}}
\newcommand{\bbpgt}[3]{\ensuremath{\hat{g}_{#1,#2}^{#3}}}
\newcommand{\setbbpgtmatched}[2]{\ensuremath{\hat{\mathtt{G}}_{#1,#2}^{\textrm{m}}}}

\def\setbbunmatched{\bar{\mathtt{D}}}
\def\bbunmatched{\bar{d}}

\newcommand{\dtconf}[1]{\ensuremath{\mathcal{S}_{\textrm{det}}(#1)}}
\newcommand{\dtcls}[1]{\ensuremath{\textrm{c}(#1)}}

\def\probdt{\ensuremath{\mathbf{p}}}

\def\probdte{\ensuremath{p}}

\def\pgtscoremin{\kappa_{\textrm{bg}}}
\def\pgtscoreign{\kappa_{\textrm{ignore}}}

\newcommand{\coco}{COCO\xspace}
\newcommand{\voc}{VOC\xspace}
\newcommand{\kitti}{KITTI\xspace}

\newcommand{\widerface}{WiderFace\xspace}

\newcommand{\lisasigns}{LISA-Signs\xspace}

\newcommand{\sunrgbd}{SUN-RGBD\xspace}

\usepackage{graphicx}

\usepackage{tikz}
\usepackage{comment} 
\usepackage{amsmath,amssymb} 
\usepackage{color}

\usepackage[pdftex,pagebackref,colorlinks,citecolor=blue,linkcolor=blue,urlcolor=blue]{hyperref}

\begin{document}
\pagestyle{headings}
\mainmatter
\def\ECCVSubNumber{2113}  

\title{Object Detection with a Unified Label Space from Multiple Datasets} 

\titlerunning{Object Detection with a Unified Label Space from Multiple Datasets}
%
\author{Xiangyun Zhao$^1$, Samuel Schulter$^2$, Gaurav Sharma$^2$, Yi-Hsuan Tsai$^2$, \\
Manmohan Chandraker$^{2,3}$, Ying Wu$^1$}
%
\authorrunning{X. Zhao, S. Schulter, G. Sharma, Y.-H. Tsai, M. Chandraker, Y. Wu}
%
\institute{$^1$Northwestern University \hspace{5mm} $^2$NEC Labs America \hspace{5mm} $^3$UC San Diego 
}
\maketitle

\begin{abstract}

Given multiple datasets with different label spaces, the goal of this work is to train a single object detector predicting over the union of all the label spaces.
The practical benefits of such an object detector are obvious and significant---application-relevant categories can be picked and merged form arbitrary existing datasets.
However, na\"ive merging of datasets is not possible in this case, due to inconsistent object annotations.
Consider an object category like faces that is annotated in one dataset, but is not annotated in another dataset, although the object itself appears in the latter's images.
Some categories, like face here, would thus be considered foreground in one dataset, but background in another.
To address this challenge, we design a framework which works with such partial annotations, and we exploit a pseudo labeling approach that we adapt for our specific case.
We propose loss functions that carefully integrate partial but correct annotations with complementary but noisy pseudo labels.
Evaluation in the proposed novel setting requires full annotation on the test set.
We collect the required annotations\footnote{Project page: \url{http://www.nec-labs.com/~mas/UniDet} \\ This work was part of XZ's internship at NEC Labs America.} and define a new challenging experimental setup for this task based on existing public datasets.
We show improved performances compared to competitive baselines and appropriate adaptations of existing work.

\end{abstract}


\section{Introduction}
\label{sec:intro}

\begin{figure}%
    \centering%
    \begin{subfigure}{0.46\textwidth}\centering%
        \includegraphics[width=0.95\textwidth]{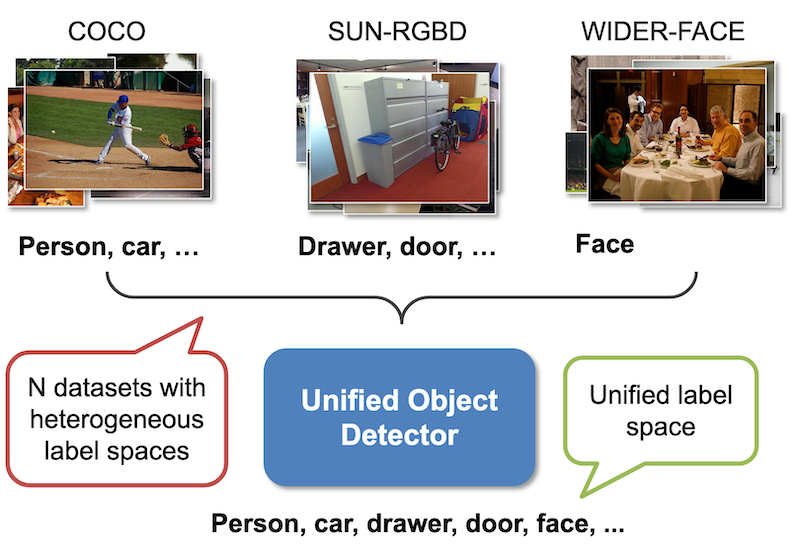}
        \caption{}
        \label{fig:teaser}
    \end{subfigure}
    \hspace{0.10cm}
    \begin{subfigure}{0.46\textwidth}\centering%
        \includegraphics[width=0.95\textwidth]{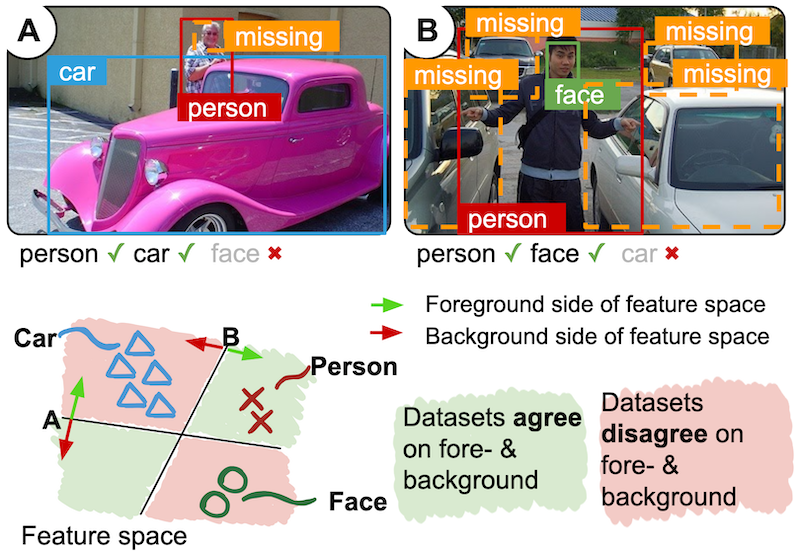}%
        \caption{}%
        \label{fig:background_issue}%
    \end{subfigure}
    \caption{\textbf{(a)} We train a single object detector from multiple datasets with heterogeneous label spaces.  In contrast to prior work~\cite{Wang_2019_CVPR}, our model unifies the label spaces of all datasets. \textbf{(b)} Illustration of the ambiguity of background in object detection when training from multiple datasets with different label spaces.  Here, only ``person'' is consistent \wrt both datasets but ``car'' and ``face'' are missing in the other one, respectively.  Na\"ive combination of the datasets leads to wrong training signals}
    \label{fig:teaser_and_background_issue}
\end{figure}

Object detection has made tremendous progress in recent years to become a powerful computer vision tool~\cite{He_2017_ICCV,Ren_2015_NIPS,Singh_2018_CVPR,zhao2018pseudo}.
This is driven by the availability of large-scale datasets with bounding box annotations.
However, obtaining such data is costly and time-consuming.
While multiple publicly available datasets with annotations for various categories already exist, their label spaces are mostly different, which makes a na\"ive combination of the data impossible \cite{Gupta_2019_ICCV,Kuznetsova_2018_arxiv,Lin_2014_ECCV}.
But such a unification could be crucial for applications that require categories not labeled in some of the datasets, to avoid a costly annotation process.
We address this problem and {\em propose to unify  heterogeneous label spaces across multiple datasets}.
We show how to train a single object detector capable of detecting over the union of all training label spaces on a given test image (see Fig.~\ref{fig:teaser}).


Unifying different label spaces is not straightforward for object detection.
A key challenge is the {\em ambiguity in the definition of the ``background'' category}.
Recall that image regions in object detection datasets may contain objects without annotation because the object's category is not part of the label space.
These image regions are then considered part of the background.
However, their object category may be annotated in a different dataset.
For instance, the label space of the \coco dataset~\cite{Lin_2014_ECCV} has 80 categories but it does not contain ``human faces'', which is present in many of the \coco images, and also annotated in other datasets like~\cite{Yang_2016_CVPR}.
A pair of images illustrate the point in Fig.~\ref{fig:background_issue}.
Hence, images and annotations of different datasets can not be trivially combined while preserving the annotation consistency \wrt the union of all their label spaces.

A straightforward attempt to handle the ambiguity of the label space is exhaustive manual annotation, which is very costly in terms of both time and money.
Recent works have tried to address a similar problem of expanding the label space of object detectors~\cite{Redmon_2017_CVPR,Singh_2018_CVPRb,Wang_2019_CVPR}, which we discuss in Sec.~\ref{sec:related_work}.
However, none of them can truly unify multiple label spaces and are, thus, not applicable in our setting.
In contrast, {\em we propose to handle the ambiguity directly and unify the label spaces in a single detector}.
At each training iteration, we sample images from a single dataset along with the corresponding ground truth.
Predicted bounding boxes of the model need to be associated with ground truth in order to compute a loss for object detectors.
While positive matches with the ground truth can be assigned a category and a regression target, unmatched boxes, which normally would be assigned the background category, now become ambiguous because of the incomplete label space (see Fig.~\ref{fig:background_issue} and Sec.~\ref{sec:unify_labelspace}).
To resolve this ambiguity, we leverage a pseudo labeling paradigm, where dataset-specific detectors are trained on individual datasets, thus not suffering the ambiguity, which are then applied on other datasets that require additional annotation.
We then propose in Sec.~\ref{sec:resolving_label_space_ambiguity} a novel association procedure and loss function that carefully integrate this potentially noisy pseudo ground truth with the actual accurate annotations for the current dataset.

Our proposed task naturally suggests a novel setting for object detection (Sec.~\ref{sec:eval_of_unified_detector}):
Given $\numds$ datasets with heterogeneous label spaces, train a detector capable of predicting over the union of all label spaces, which is evaluated on a test set equipped with annotations for all categories.
For this purpose, we choose four existing and challenging datasets (\coco~\cite{Lin_2014_ECCV}, \sunrgbd~\cite{Song_2015_CVPR}, \voc~\cite{Everingham_2010_IJCV} and \lisasigns~\cite{Mogelmose_2012_TITS}), mix their respective validation/test sets and collect novel annotations for missing categories.
Our results in Sec.~\ref{sec:experiments} show that the proposed training algorithm successfully unifies the label spaces and outperforms competitive baselines based on prior work on this practically relevant and challenging novel task.


\section{Related Work}
\label{sec:related_work}

%

The ultimate goal of our work is to build an object detector that expands its label space beyond what is annotated in a single available training dataset.  We propose an algorithmic approach to address this problem, which we contrast to related prior art and alternative approaches in this section.

\paragraph{Manual annotation:}
The most obvious attempt to expand the label space of an object detector is by manually annotating the desired categories.
New datasets are routinely proposed~\cite{Gupta_2019_ICCV,Kuznetsova_2018_arxiv,Lambert_2020_CVPR}, but these come at high cost in terms of both time and money.
Attempts to reduce these costs have been proposed either by interactive human-in-the-loop approaches~\cite{Russakovsky_2015_CVPR,Yao_2012_CVPR} or by relying on cheaper forms of annotations and developing corresponding algorithms~\cite{Bilen_2016_CVPR,Papadopoulos_2017_CVPR,WangZian_2019_CVPR}.
In the former attempt, one still needs to revisit every single image of the existing dataset for any new category added.
The latter attempt is more promising~\cite{Bilen_2016_CVPR,Kantorov_2016_ECCV,Wan_2019_CVPR,Wan_2018_CVPR}. However, it cannot compete with fully-supervised approaches so far.
In contrast, our work shows how to combine multiple datasets with bounding box annotations for heterogeneous sets of categories.


\paragraph{Object detection with bounding box and image-level annotations:}
Several works try to leverage the combination of object detection datasets and image classification datasets~\cite{Redmon_2017_CVPR,Uijlings_2018_CVPR,Yang_2019_ICCV}.  Detection datasets typically annotate coarse-grained categories with bounding boxes~\cite{Everingham_2010_IJCV,Lin_2014_ECCV}, while classification datasets exist with fine-grained category annotation but only on the image level~\cite{Deng_2009_CVPR}.  The main assumption in these works is that there are certain semantic and visual relationships between the categories with bounding box annotation and the ones with only image-level annotation.  Mostly, the fine-grained categories are a sub-category of the coarse-grained categories.  However, this assumption is a limiting factor because categories without clear visual or semantic relations are hard to combine, \eg, ``person'' and ``car''.  Our proposed approach assumes bounding box-annotated datasets as input, but has the capability to combine any categories regardless of their visual or semantic relationship.


\paragraph{Universal representations:}
The recently proposed work by Wang~\etal~\cite{Wang_2019_CVPR} on universal object detection has a similar goal as our work.
Inspired by a recent trend towards universal representations~\cite{Bilen_2017_TR,Kalluri_2019_ICCV,Rebuffi_2017_NIPS}, Wang~\etal propose to train a single detector from multiple datasets annotated with bounding boxes in a multi-task setup, which we review in more detail in Sec.~\ref{sec:preliminaries}.
However, the categories in their design are not actually unified, but kept separate.  At test time, the object detector knows from which dataset each test image comes from and thus makes predictions in the corresponding label space. In contrast, we propose to truly unify the label spaces from all training datasets with a more efficient network design.
Moreover, our experimental setup generalizes the one from~\cite{Wang_2019_CVPR} as it requests from the detector to predict all categories (unified label space) for any given test image.  This is a more challenging and realistic setup since there is no need to constrain each test image to the label space of a single training dataset.

\paragraph{Domain adaptation:}
Only a few methods have been proposed on domain adaptation for object detection~\cite{Chen_2018_CVPR,Hsu_2020_WACV,Inoue_2018_CVPR}, but they are certainly related as training from multiple datasets is naturally confronted with domain gaps.
While this is an interesting direction to explore, it is orthogonal to our contributions because none of these works address the problem of mismatched label spaces across datasets.

\paragraph{Learning from partial annotation:}
The most relevant work in this context is from Cour~\etal~\cite{Cour_2011_JMLR}, who propose loss functions for linear SVMs.
Others have tried to use pseudo labeling to address this problem~\cite{Feng_2019_AAAI}.
Most of these works, however, focus on plain classification problems, while we adapt some of these concepts to our specific task for object detection with deep networks in Sec.~\ref{sec:method}.


\section{Training with Heterogeneous Label Spaces}
\label{sec:method}

\subsection{Preliminaries}
\label{sec:preliminaries}

\paragraph{Notation:}
We work with $\numds$ datasets, $\ds{1}, \ds{2}, \ldots, \ds{\numds}$ and corresponding label spaces $\lblspc{1}, \lblspc{2}, \ldots, \lblspc{\numds}$, each $\lblspc{i}$ being a set of categories specific to dataset $i$.  In general, we do not constrain the label spaces to be equal, \ie, $\lblspc{i} \neq \lblspc{j}$ for $ i \neq j$ and we allow common categories, \ie, $\lblspc{i} \cap \lblspc{j} \neq \emptyset$ for $ i \neq j$, where $\emptyset$ is the empty set.  For example, the label spaces of \coco, \voc and \kitti are different but all of them contain the category ``person''.  We also include the special category ``background'', $b_i$, in every dataset $\ds{i}$ making the complete label space for $\ds{i}$: $\lblspc{i}\cup b_i$.  However, the definition of individual ``background'' categories $b_i$ is different.  Thus, merging the dataset-specific background is the main challenge we address below (see also Sec.~\ref{sec:intro} and Fig.~\ref{fig:background_issue}).

Dataset $\ds{i}$ contains $\dssize{i}$ images $\imgi{i}{j}$, with $j = 1, \ldots, \dssize{i}$, and their corresponding sets of ground truth annotations $\setbbgt{i}{j} = \{\bbgt{i}{j}{k} = (x_1, y_1, x_2, y_2, c)^k, k = 1, \ldots, \cardinality{\setbbgt{i}{j}}\}$.  Each ground truth annotation corresponds to an object present in the image and contains its bounding box coordinates, in image space, and category label $c \in \lblspc{i}$.  Any region of an image $\imgi{i}{j}$ which is not covered by any bounding box in $\setbbgt{i}{j}$ is considered to be in the background category $b_i$ for the respective dataset.

\paragraph{Object detection framework:}
We use Faster-RCNN~\cite{Ren_2015_NIPS} together with the Feature Pyramid Network (FPN)~\cite{Lin_2017_CVPR} as our base object detector.  Faster-RCNN takes as input an image and extracts convolutional features with a ``backbone'' network.  It then uses a region proposal network (RPN) which predicts a set of bounding boxes that describe potential (category-agnostic) objects.  Further, it has a region classification network (RCN) which classifies each proposal into the defined label space and also refines the localization.  See~\cite{Lin_2017_CVPR,Ren_2015_NIPS} for details.

\begin{figure}[!t]
  \begin{center}
  \includegraphics[width=0.6\textwidth]{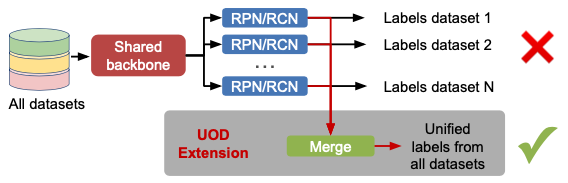}
  \end{center}
  \caption{Design of ``universal'' object detection~\cite{Wang_2019_CVPR} and our extension to unify the label spaces (UOD+Merge)}
  \label{fig:uod_archs}
\end{figure}


\paragraph{Universal object detection baseline:}
Wang~\etal~\cite{Wang_2019_CVPR} recently proposed a ``universal'' object detector (UOD) that is also trained from multiple datasets.
This is done by equipping the model with multiple RPN/RCN heads, one for each dataset, and train the model similar to a multi-task setup where each dataset can be seen as one task.
While this bypasses the ambiguity in the label spaces, at test-time, UOD needs to know what dataset each test image comes from and can thus activate the corresponding RPN/RCN head.
This is not directly applicable in our more realistic setting, because we request to detect all categories from the training datasets on any test image.
However, we introduce a simple extension to alleviate this problem, UOD+merge, which acts as a baseline to our proposed approach.
It runs all RPN/RCN detector heads on a test image and merges semantically equal categories via an additional non-maximum suppression step (see Fig.~\ref{fig:uod_archs}).

\subsection{Unifying label spaces with a single detector}
\label{sec:unify_labelspace}

The main goal of this work is to train a single detector with the union set of all label spaces $\lblspc{\cup} = \lblspc{1}\cup \lblspc{2} \ldots \cup \lblspc{\numds}$ such that given a new image $\img$, all categories can be detected.
This unified label space implicitly also defines a new and unique ``background'' category $b_{\cup}$, which is different to all other $b_i$.
The intuitive benefit of a single detector over per-dataset detector heads, as in our UOD+merge baseline, is lower computational costs.
The potentially bigger benefit, though, comes from better model parameters which are updated with data from all datasets, although having incomplete annotations.

The corresponding neural network architecture is simple:  We use a single backbone and a single detector head with the number of categories equal to $\cardinality{\lblspc{\cup}} + 1$, where $+1$ is for the background category $b_{\cup}$.
We then follow the standard training procedure of object detectors~\cite{Ren_2015_NIPS} by taking an image as input, predicting bounding boxes\footnote{
While predicted boxes can either be anchors (RPN) or proposals (RCN), we only apply our approach on RCN.  We did not observe gains when applied on RPN.
}, and assigning them to ground truth annotations to define a loss.
An association happens if predicted bounding boxes and ground truth sufficiently overlap in terms of intersection-over-union (IoU).
Successfully associated bounding boxes get assigned the category and regression target of the corresponding ground truth box.
The set of all remaining bounding boxes, which we denote as $\setbbunmatched$, would be assigned the background category (without any regression target) in a standard detection setup where ground truth is complete.
However, in the proposed setting, where the ground truth label space $\lblspc{i}$ of dataset $i$ is incomplete \wrt the unified label space $\lblspc{\cup}$, the remaining bounding boxes $\setbbunmatched$ cannot simply be assigned the background class $b_{\cup}$ because it could also belong to a category from another dataset $\lblspc{\cup} \setminus \lblspc{i}$.
To illustrate this case again, consider an object of any category $c \notin \lblspc{1}$, which will be treated as part of the background $b_1$, even though it may be present in images of $\ds{1}$.  However, this category may be annotated in another dataset $\ds{2}$ and, hence, not be part of background $b_{\cup}$.  As an example, ``human faces" are not part of the label space of the \coco dataset~\cite{Lin_2014_ECCV}, but are certainly present in \coco images and other datasets exist with such annotations, like  \widerface~\cite{Yang_2016_CVPR}.

\subsection{A loss function to deal with the ambiguous label spaces}
\label{sec:resolving_label_space_ambiguity_with_partial_anno_losses}

The problem we have to address for training a unified detector is the ambiguity of predicted bounding boxes $\setbbunmatched$ that are not associated with any ground truth of the given image from dataset $i$.
These bounding boxes either belong to a category not in the label space of the current image, \ie, $\lblspc{\cup} \setminus \lblspc{i}$, or truly are part of the unified background $b_{\cup}$.

This problem can be thought of as learning with partial annotation~\cite{Cour_2011_JMLR}, where a given sample can belong to a subset of the actual label space but the actual category is unknown.
The only constraint on the true label of the ambiguous detections $\setbbunmatched$, without additional information, is that it \emph{does not} belong to any label in $\lblspc{i}$, because these categories were annotated for the given image $\imgi{i}{j}$.
Thus, the underlying ground truth category belongs to any of $\lblspc{*} = (\lblspc{\cup} \setminus \lblspc{i}) \cup b_{\cup}$.
As suggested in~\cite{Cour_2011_JMLR}, we use this fact to design a classification loss function for ambiguous detections $\setbbunmatched$ as
\begin{equation}
  \label{eq:loss_uninformed_a}
  \mathcal{L}^{-}_{\textrm{sum}}(\probdt, \lblspc{*}) = - \log\left( \sum_{c \in \lblspc{*}} p_c \right) \;,
\end{equation}
where $\probdt = [\probdte_1, \probdte_2, \ldots] \in \realspace^{\cardinality{\lblspc{\cup}}+1}$ is the probability distribution for a predicted bounding box over the unified label space $\lblspc{\cup}$ and background, with $\sum_{c=1}^{\cardinality{\lblspc{\cup}}+1} \probdte_c = 1$.
The loss function~\eqref{eq:loss_uninformed_a} essentially is a cross-entropy loss on the sum of the ambiguous categories and can be interpreted as merging all ambiguities into one category.
An alternative is to replace the sum inside the log of \eqref{eq:loss_uninformed_a} with a max~\cite{Cour_2011_JMLR}.
An extension to~\eqref{eq:loss_uninformed_a} would be to add a minimum-entropy regularization to encourage selectivity instead of spreading probability mass over multiple categories.

While this sounds intuitive and we also use this loss as a baseline in our experiments, it has two flaws:
First, one category, the background $b_{\cup}$, is never among the certainly correct classes but always ambiguous, which may lead to problems during learning.
Second, this loss only considers the ambiguity but {\em does not resolve it}, which can be done with pseudo ground truth as described next.

\subsection{Resolving the label space ambiguity with pseudo labeling}
\label{sec:resolving_label_space_ambiguity}

\begin{figure}[t]\centering
    \includegraphics[width=1.0\textwidth]{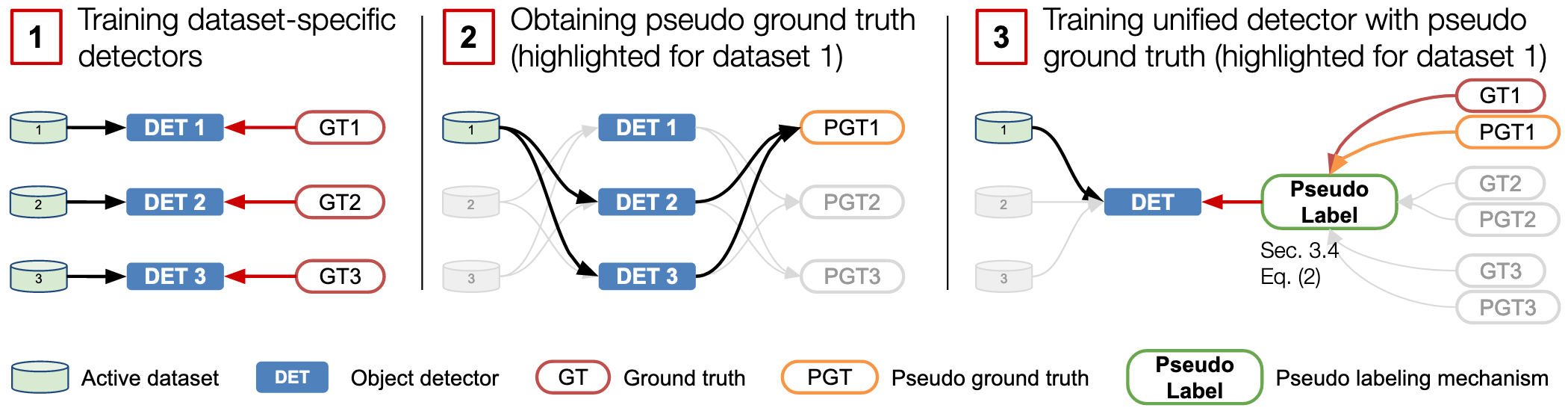}
    \caption{Overview of our pseudo labeling approach for training a single object detector with a unified label space from multiple datasets: {\bf (1)} We train dataset-specific detectors which do not suffer any label ambiguity. Although we draw three separate detectors, we share the weights of their backbones to encourage better performance across the different datasets.
    {\bf (2)} We generate pseudo ground truth for each dataset that completes the given ground truth with bounding boxes for categories not present in the current label space. Thus, images from dataset $i$ are put through detectors trained on other datasets $j \neq i$.  The figure highlights the generation of pseudo ground truth for the first dataset.
    {\bf (3)} We train a unified detector from all datasets on a unified label space $\lblspc{\cup}$.  The figure illustrates one training iteration with an image from the first dataset.  The pseudo labeling module generates the training signal for the detector given true ground truth (GT) and pseudo ground truth (PGT) from that dataset
    }
    \label{fig:pseudo_labeling_overview}
\end{figure}

Although the loss function~\eqref{eq:loss_uninformed_a} in the previous section encodes that one of the ambiguous categories in $\lblspc{*}$ is correct, it does not use any prior on the categories.
In this section, we propose to adopt a pseudo-labeling approach to impose such a prior by estimating missing annotations.
Fig.~\ref{fig:pseudo_labeling_overview} gives an overview.

\paragraph{Dataset-specific detectors:}
These are detectors trained on a single dataset $\ds{i}$ and have no ambiguity in their label space.
We use them to augment incomplete annotations on images from different datasets $\ds{j}$ with different label spaces $\lblspc{j}$ ($j \neq i$).
We use $\numds$ dataset-specific detectors for each of the $\numds$ datasets, which can be trained in different ways.
The obvious option is to independently train $\numds$ detectors, where each of them has its separate feature extraction backbone.
Another option is to use a UOD~\cite{Wang_2019_CVPR} like network where a shared backbone is followed by $\numds$ separate detector heads.

Since dataset-specific detectors are trained on one dataset $\ds{j}$ but applied to another one $\ds{i}$ to augment the annotations, the domain gap between the datasets should be considered.
Among the two training options above, we expect the latter to be relatively better since part of the network is shared across domains.
Another option for training dataset-specific detectors could be to leverage domain adaptation techniques, \eg \cite{Chen_2018_CVPR,Hsu_2020_WACV,Inoue_2018_CVPR}, which we expect to boost performance, but are orthogonal to our contributions and we leave this for future work.

Finally, we note that dataset-specific detectors are only used at train time, but are not required at inference time for our proposed unified object detector.

\paragraph{Training with pseudo labels:}
During training of our unified detector, each mini-batch contains data only from a single dataset, which enables efficient gradient computation.  Suppose we have an image $\imgi{i}{j}$ of dataset $\ds{i}$ with label space $\lblspc{i}$, its ground truth for categories in $\lblspc{i}$ are available but those in $\lblspc{*} = (\lblspc{\cup} \setminus \lblspc{i}) \cup b_{\cup}$ are not.  To remedy this, we run all dataset-specific detectors, hence covering $\lblspc{*}$, to obtain a set of bounding boxes $\setbbpgt{*}{j}$ of label space $\lblspc{*}$ for image $j$, in which we consider these to be the pseudo ground truth.  Each pseudo ground truth box $\bbpgt{*}{j}{k}$ has a detection score $\dtconf{\bbpgt{*}{j}{k}}$ associated with it. In the next paragraph, we describe a loss function that leverages this pseudo ground truth as a strong prior to resolve the label space ambiguity.

\paragraph{Loss function with pseudo labels:}
Pseudo ground truth needs to be used carefully during training because it contains noise owing to (a) the domain gap between datasets and (b) the errors from the detectors. We propose a robust loss function and matching strategy as follows. Given a set of unmatched detections $\setbbunmatched$ and all pseudo ground truth boxes $\setbbpgt{*}{j}$, we first compute the IoU similarity $s_{l,k}$ between $\bbunmatched_l \in \setbbunmatched$ and $\bbpgt{*}{j}{k} \in \setbbpgt{*}{j}$ for all $l$ and $k$.  In contrast to standard object detectors, where the closest ground truth for each anchor/proposal is chosen, we keep all pseudo ground truth boxes with sufficiently high IoU similarity (\ie, $s_{l,k} > \tau$).  We keep multiple matches with pseudo ground truth to counter the uncertainty of pseudo labeling and average out the potential noise.  Suppose for each box $\bbunmatched_l$ (not matched to $\setbbgt{i}{j}$), we have a set of matched pseudo ground truth boxes $\setbbpgtmatched{*}{j} = \{ \bbpgt{*}{j}{k} | s_{l,k} > \tau \wedge \dtconf{\bbpgt{*}{j}{k}} > \pgtscoremin \}$, where $\dtconf{\cdot}$ is the detection score. The threshold $\pgtscoremin$ defines a minimum score for a detection to be considered as pseudo ground truth class, and anything below it is considered as background $b_\cup$.
If $\setbbpgtmatched{*}{j}$ is empty, we assume the ground truth label for $\bbunmatched_l$ is ``background'' and use a standard cross-entropy loss.  Otherwise, we employ the following loss on the predicted class distribution $\probdt_l$ of $\bbunmatched_l$:
\begin{equation}
  \mathcal{L}_{\textrm{P}}(\probdt_l) = \frac{1}{Z} \sum_k \Gamma(\dtconf{\bbpgt{*}{j}{k}}) \cdot CE(\probdt_l, \dtcls{\bbpgt{*}{j}{k}}) \;,
  \label{eq:pseudo_gt_loss_cls}
\end{equation}
where $\bbpgt{*}{j}{k} \in \setbbpgtmatched{*}{j}$ and $\dtcls{\bbpgt{*}{j}{k}}$ is the category of $\bbpgt{*}{j}{k}$.
The loss is the sum over the matched pseudo ground truth boxes, weighted by $\Gamma(\dtconf{\bbpgt{*}{j}{k}}) : \realspace \rightarrow \realspace$, which decides the importance of the pseudo ground truth.  It is normalized by $Z = \max \left( \sum_k \Gamma(\dtconf{\bbpgt{*}{j}{k}}), \epsilon \right)$, where $\epsilon$ is a small constant to prevent division-by-zero in case all $\bbpgt{*}{j}{k}$ have weight $0$. There could be many different ways to define such a weighting function $\Gamma(\cdot)$, and we analyze various choices in our experiments. One important aspect we find to be crucial for the success of pseudo labeling is to ignore uncertain detections and not mark them as pseudo labels. Soft weighting by the score itself, \ie $\Gamma(x)=x$ as the identity function, is one possible instantiation of such a weighting function. Another possibility is a hard threshold, \ie $\Gamma(x; \pgtscoreign) = 1$ if $x>\pgtscoreign, 0$ otherwise. In the second case, the boxes with scores between $\pgtscoremin$ and $\pgtscoreign$ are ignored and do not contribute to the loss at all, analogous to common detectors like Faster-RCNN~\cite{Ren_2015_NIPS}.

Note that~\eqref{eq:pseudo_gt_loss_cls} only defines the classification loss based on pseudo ground truth.  We can also add the corresponding regression loss as in standard detectors like~\cite{Girshick_2015_ICCV,Ren_2015_NIPS}, which we evaluate in our experiments.

\subsection{Evaluating a unified object detector}
\label{sec:eval_of_unified_detector}

Standard object detection evaluation has typically been limited to a single dataset~\cite{He_2017_ICCV,Ren_2015_NIPS}.
We propose a more challenging and arguably more realistic experimental setup, where we take $\numds$ standard datasets with their corresponding train/val splits and {\em different} label spaces.
A unified detector takes the training sets of these $\numds$ datasets as input and, at inference time, it must be able to predict the union of all training categories on any unseen test image.

For the final evaluation, we mix the validation/test sets from each dataset together into a single larger set of images.
The trained model {\em does not} know from which dataset the images come from, which differs from the setting in~\cite{Wang_2019_CVPR}.
The detector {\em predicts all categories in all of the images}.
Such a setup requires additional annotations for evaluating the test performance.
Hence, we collected the annotations for the missing categories of the unified label space for all our test images.
Please see the supplementary material for details.


\section{Experiments}
\label{sec:experiments}


We now show results on the experimental setup proposed in Sec.~\ref{sec:eval_of_unified_detector}, with various datasets.
We evaluate the unified object detectors from Sec.~\ref{sec:method} in this novel setting, analyze their different variants, and compare with competitive baselines.

\paragraph{Datasets:}
Recall that our goal is to train from $\numds$ datasets with different label spaces, while the evaluation requires predicting over the union of all training categories on any given image. We use publicly available datasets directly for training.
Tab.~\ref{tbl:exp_setup_settings} summarizes our dataset choices.
We define several specific combinations of datasets that we believe cover a diverse set of category combinations, and one additional setup, that we use for an ablation study.
Details for each setting are given in the corresponding sections below.
Note that we do not include the recently proposed setting from Wang~\etal~\cite{Wang_2019_CVPR} with 11 datasets for two main reasons: First, the combinatorial problem of annotating missing categories in 11 datasets renders it costly. Second, several datasets in the setup of~\cite{Wang_2019_CVPR} are completely dissimilar, in a sense that object categories are disjoint from the rest, \eg, medical images \vs usual RGB images, which defeats the purpose of what we want to demonstrate in this paper. 

\paragraph{Implementation details:}
We implement all our models with PyTorch\footnote{We use the following code base as the starting point for our implementation: \url{https://github.com/facebookresearch/maskrcnn-benchmark}}.
The backbone is ResNet-50 with FPN~\cite{Lin_2017_CVPR}, which is pretrained from ImageNet~\cite{Deng_2009_CVPR}.
For each experiment, the network is trained for 50,000 iterations with a learning rate of 0.002.
For our unified detector with pseudo labeling, we initialize the backbone with the weights from the ImageNet pretrained model.
For the weighting function in~\eqref{eq:pseudo_gt_loss_cls}, our default choice is the hard-thresholding described in Sec.~\ref{sec:resolving_label_space_ambiguity}.
We follow the standard detection evaluation as in~\cite{Wang_2019_CVPR} and report mAP$^{0.5}$.

\begin{table}[!t]\centering
\caption{Different combinations of datasets that we use in our experimental setup. For each setting, a detector needs to be trained from all datasets predicting the union of their respective label spaces}
\begin{tabular}{ l | l }
\toprule
Setting  & Datasets                  \\
\midrule
A        & \voc~\cite{Everingham_2010_IJCV} + \sunrgbd~\cite{Song_2015_CVPR} + \lisasigns~\cite{Mogelmose_2012_TITS} \\
B        & \voc~\cite{Everingham_2010_IJCV} + \coco~\cite{Lin_2014_ECCV} (w/o \voc categories) + \sunrgbd~\cite{Song_2015_CVPR} \\
Ablation & \voc~\cite{Everingham_2010_IJCV} + \coco~\cite{Lin_2014_ECCV} (w/o \voc categories) \\
\bottomrule
\end{tabular}
\label{tbl:exp_setup_settings}
\end{table}

\subsection{Ablation study}
\label{sec:ablation_study}

To better understand the importance of the individual components and variations in our proposed method, we first conduct an analysis in a controlled setting.

\paragraph{Experimental setup:}
We use a special setup where we intentionally remove certain categories, from one dataset, that are available in another dataset.
Specifically, we use \voc~\cite{Everingham_2010_IJCV} (2007) and \coco~\cite{Lin_2014_ECCV} (2017), with (annotations of) \voc categories removed from \coco.
In this way, no additional annotation is required for evaluation and we can even monitor pseudo labeling performance on the training set.
We evaluate on the combination of both datasets as described in Sec.~\ref{sec:eval_of_unified_detector} (MIX), but also on the individual ones (\coco and \voc), as well as specifically on the 20 \voc categories on the \coco dataset (V-on-C).

\paragraph{Baselines:}
We compare with the following baselines:
\begin{itemize}
\item {\bf Individual detectors:} The most apparent baseline is to train individual detectors on each of the datasets and run all of them on a new test image. As with our UOD+merge baseline, we apply a merging step as described in Sec.~\ref{sec:preliminaries}. Besides the obvious drawback of high runtime, the domain gap between datasets can lead to issues.
\item {\bf Universal detectors (UOD)}: Our proposed extension of~\cite{Wang_2019_CVPR}, UOD+Merge, described in Sec.~\ref{sec:preliminaries} and Fig.~\ref{fig:uod_archs}.
\item {\bf Unified detector ignoring label space ambiguity}: This baseline (``Unify w/o Pseudo-Labeling'' in the tables) trains a single detector head with the unified label space $\lblspc{\cup}$ but treats all unmatched detections $\setbbunmatched$ as part of the unified background $b_{\cup}$. It thus ignores the ambiguity in the label spaces.
\item {\bf Partial annotation losses}: This baseline implements the unified detector head and tries to resolve the label space ambiguity with the loss function~\eqref{eq:loss_uninformed_a} designed for partial annotations~\cite{Cour_2011_JMLR}. As mentioned in~\ref{sec:resolving_label_space_ambiguity_with_partial_anno_losses}, we evaluate alternatives to~\eqref{eq:loss_uninformed_a}: (i) We add a minimum entropy regularization to~\eqref{eq:loss_uninformed_a} (sum + ent) and (ii) we replace the sum in~\eqref{eq:loss_uninformed_a} with a $\max$ function as suggested in~\cite{Cour_2011_JMLR}.
\item {\bf Pseudo labeling}: We analyze different variants of the pseudo-labeling strategy. Instead of hard-thresholding in the weighting function $\Gamma(\cdot)$ of the loss~\eqref{eq:pseudo_gt_loss_cls}, which is the default, we also evaluate soft-thresholding (soft-thresh) as described in Sec.~\ref{sec:resolving_label_space_ambiguity} (paragraph ``Loss function with pseudo labels").  Given the inherent uncertainty in pseudo ground truth, we additionally analyze its impact on classification and regression losses by adding a variant where we remove the regression loss for every pseudo ground truth match (w/o regression).
\end{itemize}

\begin{table}[!t]\centering\small
  \caption{Results of our ablation study on the four different evaluation sets. Please see text for more details on the training sets and the baselines.}
  \begin{tabular}{ l | r r r r }
    \toprule
    Method & \rotatebox{90}{\voc} & \rotatebox{90}{\coco} & \rotatebox{90}{V-on-C} & \rotatebox{90}{MIX}  \\
    \midrule
    \midrule
    (a) Individual detectors + Merge            & \textbf{81.4} & 46.1 & \textit{62.0} & 46.9 \\
    (b) UOD~\cite{Wang_2019_CVPR} + Merge       & 81.0 & 45.6 & 61.5 & 46.1 \\
    \midrule
    \midrule
    (c) Unify w/o Pseudo-Labeling               & 79.5 & 42.6 &  36.5 & 43.7 \\
    \midrule
    (d) Partial-loss sum~\cite{Cour_2011_JMLR}~\eqref{eq:loss_uninformed_a} & 78.9 & 43.4 & 39.2 & 44.1\\
    (e) Partial-loss sum + minimum-entropy     & 79.1 & 42.9 & 38.4& 43.9 \\
    (f) Partial-loss max~\cite{Cour_2011_JMLR} & 79.9 & 43.6 &  36.1& 44.6  \\
    \midrule
    (g) Pseudo-Labeling (soft-thresh)          & 80.5 & 49.6 & 61.6 & 51.6  \\
    (h) Pseudo-Labeling (w/o regression)       & 80.9 & \textit{50.1} & 61.9 & \textit{52.0} \\
    (i) Pseudo-Labeling                        & \textit{81.3} & \textbf{50.3} & \textbf{62.2} & \textbf{52.2}  \\
    \bottomrule
  \end{tabular}
  \label{tbl:results_ablation}
\end{table}

\paragraph{Results:}
Tab.~\ref{tbl:results_ablation} compares our final pseudo labeling model (Pseudo-Labeling) with various baselines. We make several observations:

First and most importantly, we can observe a clear improvement of pseudo labeling (i) over the most relevant baseline, UOD+Merge (b).  A big improvement (46.1 to 52.2) can be observed on the MIX setting, which is arguably the most relevant one since it requests to detect the union of all training label spaces on test images from all domains.  We identified a significantly lower number of false positives of pseudo labeling (i) as one of the reasons for this big improvement.  The baselines (a-b) suffer more from the domain gap because all (a) or some parts (b) of the detector are trained only for a single dataset.  The difference between having individual detectors (a) versus a shared backbone (b) seems negligible, but (b) obviously has a better runtime. Overall, the proposed pseudo-labeling (i) is still the most efficient one in terms of number of operations.

Second, methods that ignore the label space ambiguity (c) clearly underperform compared to other methods, particularly on the sets V-on-C and MIX, where it is important to have categories from one dataset transferred to another.

Third, the loss functions for problems with partial annnotation (d-f) improve only marginally over ignoring the label space ambiguity (c).  This is expected to some degree, as we pointed out a potential flaw in Sec.~\ref{sec:resolving_label_space_ambiguity_with_partial_anno_losses}, which is distinct to the task of object detection.  The background category stays ambiguous for all examples while other categories are part of the certain classes at least every $\numds$ samples, $\numds$ being the number of datasets.

Finally, comparing different variants of pseudo labeling itself, \ie, (g-i), hard-thresholding (i) outperforms soft-thresholding (g) slightly.  Also, using a regression loss on the pseudo ground truth (i) improves compared to (h).

\paragraph{Evaluation of pseudo labels:}
Our ablation study setting allows us to evaluate the performance of pseudo labeling itself, because ground truth is available.  Specifically, we can evaluate the \voc detector on the \coco train set.  Using all pseudo detections, we get precision and recall of 0.29 and 0.7, respectively.  However, restricting the evaluation to only confident pseudo detections, we get precision and recall of 0.79 and 0.4, demonstrating the importance of thresholding (or weighing) pseudo detections based on confidence as we do in our loss~\eqref{eq:pseudo_gt_loss_cls}.


\subsection{Comparing pseudo labeling with an upper bound}
In this section, we compare our pseudo labeling approach with an upper bound that has access to the full annotations.
We split the \coco train set into two even halves, where each half sees 50\% of the images and ignores 40 (different) categories.  We train a standard detector on the original \coco train set and our pseudo detector and UOD on the splitted data set.  The standard detector (upper bound) obtains 50.2 AP50, our pseudo detector 48.4 and UOD+merge 45.7. With domain gap absent for this experiment, we see that our pseudo labeling achieves competitive results compared to the upper bound.

\subsection{Main results}
In this section, we present our main results where we compare our proposed unified detectors, (e) and (i) in Tab.~\ref{tbl:results_ablation}, with the two most relevant baselines, UOD~\cite{Wang_2019_CVPR} + Merge (b) and Partial-anno loss based on the loss from~\cite{Cour_2011_JMLR} (d).
We first provide details on the two different settings we chose for evaluation, summarized in Tab.~\ref{tbl:exp_setup_settings} before we show quantitative and qualitative results.
Note that for each of the 4 datasets in use, we collect annotations for 500 images with the unified label space.

\paragraph{Setting A} combines 3 different object detection datasets from different domains: general (\voc~\cite{Everingham_2010_IJCV}), indoor scenes (\sunrgbd~\cite{Song_2015_CVPR}) and driving scenes (\lisasigns~\cite{Mogelmose_2012_TITS}).  The datasets have 20, 18 and 4 categories, respectively, with a few overlapping ones in the label spaces. The unified label space has 38 categories in total (4 overlapping).  For the three datasets, we have a total of 1500 images for evaluation.  Note that the images from different datasets are mixed for evaluation.


\paragraph{Setting B} has three main intentions compared to setting A: First, we increase the label space with the addition of \coco~\cite{Lin_2014_ECCV}.  Second, we increase the number of categories that need to be transferred between different datasets by removing the \voc categories from \coco. Third, we use a more focused set of scenes by using the combination of general scenes (\voc and \coco) and indoor scenes (\sunrgbd). The datasets have 20, 60 and 18 categories, respectively. The unified label space contains 91 categories. Again, for the three datasets, we have a total of 1500 images, which are mixed together for evaluation.

\begin{table}[!t]\centering
    \caption{\textbf{(a)} Results on the two main settings as defined in Tab.~\ref{tbl:exp_setup_settings}. \textbf{(b)} Performance of pseudo labeling for overlapping classes. All numbers are AP50~\cite{Lin_2014_ECCV}}.
    \begin{subtable}{0.43\textwidth}
        \begin{tabular}{ l | r r}
        \toprule
        Method \textbackslash~Setting & A & B\\
        \midrule
        UOD~\cite{Wang_2019_CVPR} + Merge         & 59.3 & 43.7  \\
        Partial-anno loss~\cite{Cour_2011_JMLR}   & 58.3 & 41.2  \\
        (Ours) Min-Entropy loss                   & 58.7 & 42.9  \\
        (Ours) Pseudo-Labeling                    & 61.1 & 48.5  \\
        \bottomrule
        \end{tabular}
        \caption{}
        \label{tbl:results_all_setting}
    \end{subtable}
    \hfill
    \begin{subtable}{0.54\textwidth}
        \begin{tabular}{ l | r r r r r r}
        \toprule
        Method \textbackslash~Class & chair &	sofa &	tv & bed &	toilet & sink\\
        \midrule
        Individ (VOC)         &\multicolumn{1}{c}{68.6}	& \multicolumn{1}{c}{74.8}	&\multicolumn{1}{c}{81} & \multicolumn{1}{c}{-} &\multicolumn{1}{c}{-} &\multicolumn{1}{c}{-}  \\
        Unified (VOC)   & \multicolumn{1}{c}{62.3}&	\multicolumn{1}{c}{76.3}&	\multicolumn{1}{c}{77.6} & \multicolumn{1}{c}{-}&\multicolumn{1}{c}{-}&\multicolumn{1}{c}{-} \\
        \midrule

        Individ (COCO)               & \multicolumn{1}{c}{-}&\multicolumn{1}{c}{-}&\multicolumn{1}{c}{-}    & \multicolumn{1}{c}{55.0}&	\multicolumn{1}{c}{66.6}&	\multicolumn{1}{c}{51.5}  \\
        Unified (COCO)                 & \multicolumn{1}{c}{-}&\multicolumn{1}{c}{-}&\multicolumn{1}{c}{-}   & \multicolumn{1}{c}{56.7}&	\multicolumn{1}{c}{68.9} &	\multicolumn{1}{c}{53.7}  \\
        \midrule
        Individ (SUN)                   &\multicolumn{1}{c}{75.3}&	\multicolumn{1}{c}{68.2}&	\multicolumn{1}{c}{85.1}&	\multicolumn{1}{c}{78.1} &	\multicolumn{1}{c}{67.2} &	\multicolumn{1}{c}{78.0} \\
        Unified (SUN)                    & \multicolumn{1}{c}{75.2}&	\multicolumn{1}{c}{64.5}	&\multicolumn{1}{c}{85.5} &	\multicolumn{1}{c}{78.6} &	\multicolumn{1}{c}{66.5} &	\multicolumn{1}{c}{81}  \\
        \bottomrule
        \end{tabular}
        \caption{}
        \label{tbl:overclasses_result}
    \end{subtable}
    \label{tbl:combo}
\end{table}

\paragraph{Quantitative results:}
Tab.~\ref{tbl:results_all_setting} provides the results on the two settings described above.
We compare our proposed pseudo labeling method with three competitive baselines: UOD~\cite{Wang_2019_CVPR} + Merge, partial-annotation-loss~\cite{Cour_2011_JMLR} and minimum-entropy-loss.
In both settings (A and B from Tab.~\ref{tbl:exp_setup_settings}), we can observe the same trend.
Pseudo labeling significantly outperforms all baselines, \eg $61.1$ \vs $59.3, 58.3$ and $58.7$, respectively, on setting A, which again verifies the effectiveness of leveraging dataset-specific detectors as prior for training a unified detector.
On setting B, we also observe an improvement of $+4.8$ AP50 points with pseudo labeling over the most relevant baseline UOD+Merge.
The improvements are mostly attributed to the reduced number of false positives compared with UOD+Merge.
Similar to the ablation study in Sec.~\ref{sec:ablation_study}, we see that only using loss functions designed for dealing with partial annotations is not enough to resolve the label space ambiguity for the task of object detection.

\begin{figure}\centering
  \includegraphics[width=\textwidth]{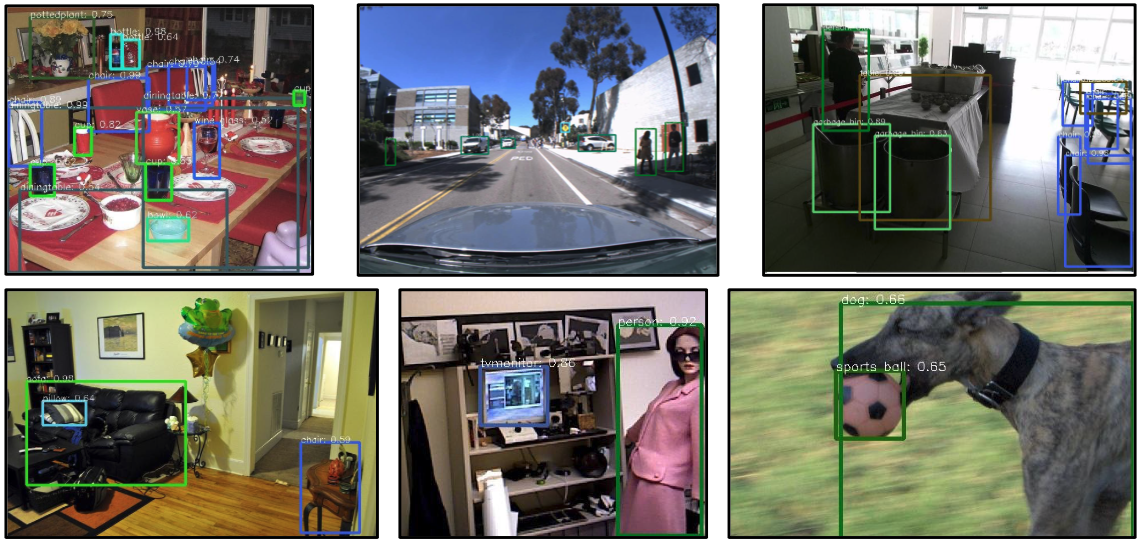}
  \caption{Some qualitative results of our proposed unified object detector}
  \label{fig:qual1}
\end{figure}

\paragraph{Performance of pseudo labeling for overlapping classes:}
Here, we specifically analyze the performance of pseudo labeling for overlapping classes among different datasets.  For setting B, we have: \voc and \sunrgbd share “chair”, “sofa” and “tv-monitor”; \coco and \sunrgbd share “bed”, “toilet” and “sink”. To analyze the difference between individual detectors and our unified detector (w/ pseudo labeling), we test on each dataset separately.  The results in Tab.~\ref{tbl:overclasses_result} show that some categories improve, likely due to increased training data via pseudo labeling, while others get worse, eventually due to domain gap.  However, we highlight that our proposed unified detector obtains much better results when tested on the mix of all data sets (with a complete label space) compared to individual detectors that may generate false positives on images from other data sets (MIX column in Tab~\ref{tbl:results_ablation}).

\paragraph{Qualitative results:}

Finally, we show a few qualitative results for the settings A and B in Fig.~\ref{fig:qual1}.
Most notably, we can observe in all examples, the unified detector is able to successfully predict categories that are originally not in the training dataset of the respective test image.
For instance, the second image in the top row in Fig.~\ref{fig:qual1} is from the \lisasigns dataset, which does not annotate the category ``person''. This category is successfully transferred from other datasets.

\section{Conclusions}

We introduce a novel setting for object detection, where multiple datasets with heterogeneous label spaces are given for training and the task is to train an object detector that can predict over the union of all categories present in the training label spaces.
This setting is challenging and more realistic than standard object detection settings~\cite{Ren_2015_NIPS}, as well as the recent ``universal'' setting~\cite{Wang_2019_CVPR}, since there can be both (i) unannotated objects of interest in training images, as well as, (ii) the original dataset of the test image is unknown.
We propose training algorithms that leverage (i) loss functions for partially annotated ground truth and (ii) pseudo labeling techniques, to build a single detector with a unified label space. We also collect new annotations on test image to enable evaluation on the different methods in this novel setting.


\subsubsection*{Acknowledgements}
This work was supported in part by National Science 
Foundation grant IIS-1619078, IIS-1815561.

\clearpage
%
%
\bibliographystyle{splncs04}
\bibliography{myshortstrings,biblist}
\end{document}